\documentclass{article}

\usepackage{microtype}
\usepackage{graphicx}
\usepackage{booktabs} 

\usepackage{hyperref}

\usepackage{amsmath}
\usepackage{mathrsfs}
\usepackage{amsfonts}
\usepackage{amsthm}
\usepackage{stmaryrd}
\usepackage{xcolor}
\usepackage{colortbl}
\usepackage{cases}
\usepackage{subcaption}
\usepackage{multirow}
\usepackage{chngcntr}
\usepackage{enumitem}
\usepackage{mathtools}
\usepackage{booktabs}

\usepackage{algorithm}
\usepackage{algorithmic}

\newcommand{\cX}{\mathcal{X}}

\newcommand{\cA}{\mathcal{A}}

\newcommand{\cT}{\mathcal{T}}

\newcommand{\cL}{\mathcal{L}}

\newcommand{\bR}{\mathbb{R}}

\DeclareMathOperator*{\expect}{{\huge \mathbb{E}}}

\DeclareMathOperator*{\argmax}{arg\,max}

\usepackage{verbatim}
\usepackage{cancel}
\usepackage{thmtools,thm-restate}

\usepackage[accepted]{icml2018}

\icmltitlerunning{Implicit Quantile Networks for Distributional Reinforcement Learning}

\begin{document}

\twocolumn[
\icmltitle{Implicit Quantile Networks for Distributional Reinforcement Learning}



\icmlsetsymbol{equal}{*}

\begin{icmlauthorlist}
\icmlauthor{Will Dabney}{equal,dm}
\icmlauthor{Georg Ostrovski}{equal,dm}
\icmlauthor{David Silver}{dm}
\icmlauthor{R\'emi Munos}{dm}
\end{icmlauthorlist}

\icmlaffiliation{dm}{DeepMind, London, UK}

\icmlcorrespondingauthor{Will Dabney}{wdabney@google.com}
\icmlcorrespondingauthor{Georg Ostrovski}{ostrovski@google.com}

\icmlkeywords{Reinforcement learning, distributional reinforcement learning, machine learning, quantile regression}

\vskip 0.3in
]



\printAffiliationsAndNotice{\icmlEqualContribution} 

\begin{abstract}
In this work, we build on recent advances in distributional reinforcement learning to give a generally applicable, flexible, and state-of-the-art distributional variant of DQN. We achieve this by using quantile regression to approximate the full quantile function for the state-action return distribution. By reparameterizing a distribution over the sample space, this yields an implicitly defined return distribution and gives rise to a large class of risk-sensitive policies. We demonstrate improved performance on the 57 Atari 2600 games in the ALE, and use our algorithm's implicitly defined distributions to study the effects of risk-sensitive policies in Atari games.
\end{abstract}

\section{Introduction}
\label{sec:introduction}

Distributional reinforcement learning  \cite{jaquette73markov,sobel82variance,white88mean,morimura2010nonparametric,c51} focuses on the intrinsic randomness of returns within the reinforcement learning (RL) framework. As the agent interacts with the environment, irreducible randomness seeps in through the stochasticity of these interactions, the approximations in the agent's representation, and even the inherently chaotic nature of physical interaction \cite{yu2016more}. Distributional RL aims to model the distribution over returns, whose mean is the traditional value function, and to use these distributions to evaluate and optimize a policy.

Any distributional RL algorithm is characterized by two aspects: the parameterization of the return distribution, and the distance metric or loss function being optimized. Together, these choices control assumptions about the random returns and how approximations will be traded off. Categorical DQN \citep[C51]{c51} combines a categorical distribution and the cross-entropy loss with the Cram\'er-minimizing projection \cite{rowland2018analysis}. For this, it assumes returns are bounded in a known range and trades off mean-preservation at the cost of overestimating variance.

C51 outperformed all previous improvements to DQN on a set of 57 Atari 2600 games in the Arcade Learning Environment \cite{bellemare13arcade}, which we refer to as the Atari-57 benchmark. Subsequently, several papers have built upon this successful combination to achieve significant improvements to the state-of-the-art in Atari-57 \cite{hessel2018rainbow,gruslys2018reactor}, and challenging continuous control tasks \cite{barthmaron2018d4pg}.

These algorithms are restricted to assigning probabilities to an a priori fixed, discrete set of possible returns. \citet{dabney2017qr} propose an alternate pair of choices, parameterizing the distribution by a uniform mixture of Diracs whose locations are adjusted using quantile regression. Their algorithm, QR-DQN, while restricted to a discrete set of quantiles, automatically adapts return quantiles to minimize the Wasserstein distance between the Bellman updated and current return distributions. This flexibility allows QR-DQN to significantly improve on C51's Atari-57 performance.

In this paper, we extend the approach of \citet{dabney2017qr}, from learning a discrete set of quantiles to learning the full quantile function, a continuous map from probabilities to returns. When combined with a base distribution, such as $U([0,1])$, this forms an implicit distribution capable of approximating any distribution over returns given sufficient network capacity. Our approach, \textit{implicit quantile networks} (IQN), is best viewed as a simple distributional generalization of the DQN algorithm \cite{mnih15nature}, and provides several benefits over QR-DQN. 

First, the approximation error for the distribution is no longer controlled by the number of quantiles output by the network, but by the size of the network itself, and the amount of training. Second, IQN can be used with as few, or as many, samples per update as desired, providing improved data efficiency with increasing number of samples per training update. Third, the implicit representation of the return distribution allows us to expand the class of policies to more fully take advantage of the learned distribution. Specifically, by taking the base distribution to be non-uniform, we expand the class of policies to $\epsilon$-greedy policies on arbitrary distortion risk measures \cite{yaari1987dual,wang1996premium}.

We begin by reviewing distributional reinforcement learning, related work, and introducing the concepts surrounding risk-sensitive RL. In subsequent sections, we introduce our proposed algorithm, IQN, and present a series of experiments using the Atari-57 benchmark, investigating the robustness and performance of IQN. Despite being a simple distributional extension to DQN, and forgoing any other improvements, IQN significantly outperforms QR-DQN and nearly matches the performance of Rainbow, which combines many orthogonal advances. 
In fact, in human-starts as well as in the hardest Atari games (where current RL agents still underperform human players) IQN improves over Rainbow.

\section{Background / Related Work}
\label{sec:background}

We consider the standard RL setting, in which the interaction of an agent and
an environment is modeled as a Markov Decision Process 
$(\mathcal{X}, \mathcal{A}, R, P, \gamma)$ \cite{puterman94markov}, 
where $\mathcal{X}$ and $\mathcal{A}$ denote the state and action spaces, 
$R$ the (state- and action-dependent) reward function,
$P(\cdot | x, a)$ the transition kernel, 
and $\gamma \in (0, 1)$ a discount factor. A policy $\pi(\cdot | x)$ maps a state to a distribution over actions.

For an agent following policy $\pi$, the discounted sum of future 
rewards is denoted by the random variable $Z^\pi(x, a) = \sum_{t=0}^\infty \gamma^t R(x_t, a_t)$, where $x_0 = x$, $a_0 = a$, $x_t \sim P(\cdot | x_{t-1}, a_{t-1})$, and $a_t \sim \pi(\cdot | x_{t})$. The action-value function is defined as
$Q^\pi(x,a) = \mathbb{E}\left[ Z^\pi(x,a)\right]$, and can be characterized by the Bellman equation 
\begin{equation*}
Q^\pi(x,a) = \mathbb{E} \left[ R(x,a) \right] + 
\gamma \mathbb{E}_{P,\pi} \left[ Q^\pi(x',a')\right]. 
\end{equation*}
The objective in RL is to find an optimal policy $\pi^*$, which maximizes $\mathbb{E}[Z^\pi]$,
i.e.~$Q^{\pi^*}(x,a) \geq Q^{\pi}(x,a)$ for all $\pi$ and all $x, a$. One approach is to find the unique fixed point $Q^* = Q^{\pi^*}$ of the Bellman optimality operator \cite{bellman57dynamic}: 
\begin{equation*}
Q(x,a) = \cT Q(x,a) := \mathbb{E}\left[ R(x,a) \right] + \gamma \mathbb{E}_{P}  \max_{a'} Q(x',a').
\end{equation*}
To this end, Q-learning \cite{watkins1989learning} iteratively improves an estimate, $Q_\theta$, of the optimal action-value function, $Q^*$, by repeatedly applying the Bellman update:
\begin{equation*}
Q_\theta(x,a) \leftarrow \mathbb{E} \left[ R(x,a) \right] + 
\gamma \mathbb{E}_{P} \left[\max_{a'} Q_\theta(x',a')\right]. 
\end{equation*}
The action-value function can be approximated by a parameterized function $Q_\theta$ (e.g.~a neural network), and trained by minimizing the squared temporal difference (TD) error,
\begin{equation*}
    \delta_t^2 = \left[ r_t + \gamma \max_{a' \in \cA} Q_\theta(x_{t+1}, a') - Q_\theta(x_t, a_t) \right]^2,
\end{equation*}
over samples $(x_t, a_t, r_t, x_{t+1})$ observed while following an $\epsilon$-greedy policy over $Q_\theta$. This policy acts greedily with respect to $Q_\theta$ with probability $1 - \epsilon$ and uniformly at random otherwise.
DQN \cite{mnih15nature} uses a convolutional neural network to parameterize $Q_\theta$ and the Q-learning algorithm to achieve human-level play on the Atari-57 benchmark.

\subsection{Distributional RL}

In distributional RL, the distribution over returns (the law of $Z^\pi$) is considered instead of the scalar value function $Q^\pi$ that is its expectation. This change in perspective has yielded new insights into the dynamics of RL \cite{azar2012sample}, and been a useful tool for analysis \cite{lattimore2012pac}. Empirically, distributional RL algorithms show improved sample complexity and final performance, as well as increased robustness to hyperparameter variation \cite{barthmaron2018d4pg}.

An analogous distributional Bellman equation of the form
\begin{equation*}
Z^\pi(x,a) \stackrel{D}{=} R(x,a) + \gamma  Z^\pi(X',A')
\end{equation*}
can be derived, where $A \stackrel{D}{=} B$ denotes that 
two random variables $A$ and $B$ have equal probability laws, and the 
random variables $X'$ and $A'$ are distributed according to $P(\cdot | x, a)$ and $\pi(\cdot | x')$, respectively.

\citet{morimura10parametric} defined the distributional Bellman operator explicitly in terms of conditional probabilities, parameterized by the mean and scale of a Gaussian or Laplace distribution, and minimized the Kullback-Leibler (KL) divergence between the Bellman target and the current estimated return distribution. However, the distributional Bellman operator is not a contraction in the KL.

As with the scalar setting, a distributional Bellman optimality operator can be defined by
\begin{equation*}
\cT Z(x,a) \stackrel{D}{:=} R(x,a) + \gamma Z(X',\argmax_{a' \in \cA} \expect Z(X', a')),
\end{equation*}
with $X'$ distributed according to $P(\cdot | x, a)$. While the distributional Bellman operator for policy evaluation is a contraction in the $p$-Wasserstein distance \cite{c51}, this no longer holds for the control case. Convergence to the optimal policy can still be established, but requires a more involved argument.

\citet{c51} parameterize the return distribution as a categorical distribution over a fixed set of equidistant points and minimize the KL divergence to the projected distributional Bellman target. Their algorithm, C51, outperformed previous DQN variants on the Atari-57 benchmark. Subsequently, \citet{hessel2018rainbow} combined C51 with enhancements such as prioritized experience replay \cite{schaul16prioritized}, $n$-step updates \cite{sutton1988learning}, and the dueling architecture \cite{wang2016dueling}, leading to the Rainbow agent, current state-of-the-art in Atari-57.

The categorical parameterization, using the projected KL loss, has also been used in recent work to improve the critic of a policy gradient algorithm, D4PG, achieving significantly improved robustness and state-of-the-art performance across a variety of continuous control tasks \cite{barthmaron2018d4pg}.

\subsection{$p$-Wasserstein Metric}

The $p$-Wasserstein metric, for $p \in [1, \infty]$, plays a key role in recent results in distributional RL \cite{c51,dabney2017qr}. It has also been a topic of increasing interest in generative modeling \cite{wgan,bousquet2017optimal,tolstikhin2017wasserstein}, because unlike the KL divergence, the Wasserstein metric inherently trades off approximate solutions with likelihoods.

The $p$-Wasserstein distance is the $L_p$ metric on inverse cumulative distribution functions (c.d.f.), also known as quantile functions \cite{muller1997integral}. For random variables $U$ and $V$ with quantile functions $F_U^{-1}$ and $F_V^{-1}$, respectively, the $p$-Wasserstein distance is given by
\begin{equation*}
    W_p(U, V) = \left( \int_0^1 |F_U^{-1}(\omega) - F_V^{-1}(\omega)|^p d\omega \right)^{1/p}.
\end{equation*}

The class of optimal transport metrics express distances between distributions in terms of the minimal cost for transporting mass to make the two distributions identical. This cost is given in terms of some metric, $c\colon \cX \times \cX \to \bR^{\geq0}$, on the underlying space $\cX$. The $p$-Wasserstein metric corresponds to $c = L_p$. We are particularly interested in the Wasserstein metrics due to the predominant use of $L_p$ spaces in mean-value reinforcement learning.

\subsection{Quantile Regression for Distributional RL}
\citet{c51} showed that the distributional Bellman operator is a contraction in the $p$-Wasserstein metric, but as the proposed algorithm did not itself minimize the Wasserstein metric, this left a theory-practice gap for distributional RL. Recently, this gap was closed, in both directions. First and most relevant to this work, \citet{dabney2017qr} proposed the use of \textit{quantile regression} for distributional RL and showed that by choosing the quantile targets suitably the resulting projected distributional Bellman operator is a contraction in the $\infty$-Wasserstein metric. Concurrently, \citet{rowland2018analysis} showed the original class of categorical algorithms are a contraction in the Cram\'er distance, the $L_2$ metric on cumulative distribution functions.

By estimating the quantile function at precisely chosen points, QR-DQN minimizes the Wasserstein distance to the distributional Bellman target \cite{dabney2017qr}. This estimation uses \textit{quantile regression}, which has been shown to converge to the true quantile function value when minimized using stochastic approximation \cite{qrbook}. 

In QR-DQN, the random return is approximated by a uniform mixture of $N$ Diracs,
\begin{equation*}
    Z_\theta(x,a) := \tfrac{1}{N} \sum_{i=1}^N \delta_{\theta_i(x,a)},
\end{equation*}
with each $\theta_i$ assigned a fixed quantile target, $\hat{\tau}_i = \frac{\tau_{i-1} + \tau_{i}}{2}$ for $1 \le i \le N$, where $\tau_i = i/N$. These quantile estimates are trained using the \citet{huber1964robust} quantile regression loss, with threshold $\kappa$,
\begin{align*}
    \rho^\kappa_\tau(\delta_{ij}) &= |\tau - \mathbb{I}{\{ \delta_{ij} < 0 \}}| \frac{\cL_\kappa(\delta_{ij})}{\kappa},\ \quad \text{with}\\
    \cL_\kappa(\delta_{ij}) &= \begin{cases}
        \frac{1}{2} \delta_{ij}^2,\quad \ &\text{if } |\delta_{ij}| \le \kappa\\
        \kappa (|\delta_{ij}| - \frac{1}{2}\kappa),\quad \ &\text{otherwise}
    \end{cases},
\end{align*}
on the pairwise TD-errors
$$\delta_{ij} = r + \gamma \theta_j(x', \pi(x')) - \theta_i(x, a).$$


At the time of this writing, QR-DQN achieves the best performance on Atari-57, human-normalized mean and median, of all agents that do not combine distributional RL, prioritized replay, and $n$-step updates \cite{dabney2017qr,hessel2018rainbow,gruslys2018reactor}.

\subsection{Risk in Reinforcement Learning}

Distributional RL algorithms have been theoretically justified for the Wasserstein and Cram\'er metrics \cite{c51,rowland2018analysis}, and learning the distribution over returns, in and of itself, empirically results in significant improvements to data efficiency, final performance, and stability \cite{c51,dabney2017qr,gruslys2018reactor,barthmaron2018d4pg}. However, in each of these recent works the policy used was based entirely on the mean of the return distribution, just as in standard reinforcement learning. A natural question arises: can we expand the class of policies using information provided by the distribution over returns (i.e.~to the class of risk-sensitive policies)? Furthermore, when would this larger policy class be beneficial?

Here, `risk' refers to the uncertainty over possible outcomes, and \emph{risk-sensitive} policies are those which depend upon more than the mean of the outcomes. At this point, it is important to highlight the difference between \emph{intrinsic uncertainty}, captured by the distribution over returns, and \emph{parametric uncertainty}, the uncertainty over the value estimate typically associated with Bayesian approaches such as PSRL \cite{osband2013more} and Kalman TD \cite{geist2010kalman}. Distributional RL seeks to capture the former, which classic approaches to risk are built upon\footnote{One exception is the recent work \cite{moerland2017efficient} towards combining both forms of uncertainty to improve exploration.}.

Expected utility theory states that if a decision policy is consistent with a particular set of four axioms regarding its choices then the decision policy behaves as though it is maximizing the expected value of some utility function $U$ \cite{von1947theory},
$$\pi(x) = \argmax_a \expect_{Z(x, a)} [U(z)].$$
This is perhaps the most pervasive notion of risk-sensitivity. A policy maximizing a linear utility function is called \emph{risk-neutral}, whereas concave or convex utility functions give rise to \emph{risk-averse} or \emph{risk-seeking} policies, respectively. Many previous studies on risk-sensitive RL adopt the utility function approach \cite{howard1972risk,marcus1997risk,maddison2017particle}.

A crucial axiom of expected utility is \textit{independence}: given random variables $X$, $Y$ and $Z$, such that $X \succ Y$ ($X$ preferred over $Y$), any mixture between $X$ and $Z$ is preferred to the same mixture between $Y$ and $Z$ \cite{von1947theory}. Stated in terms of the cumulative probability functions, $\alpha F_X + (1 - \alpha) F_Z \ge \alpha F_Y + (1 - \alpha) F_Z,\ \forall \alpha \in [0, 1]$. This axiom in particular has troubled many researchers because it is consistently violated by human behavior \cite{tversky1992advances}. The Allais paradox is a frequently used example of a decision problem where people violate the independence axiom of expected utility theory \cite{allais1990allais}.

However, as \citet{yaari1987dual} showed, this axiom can be replaced by one in terms of convex combinations of outcome values, instead of mixtures of distributions. Specifically, if as before $X \succ Y$, then for any $\alpha \in [0, 1]$ and random variable $Z$, $\alpha F_X^{-1} + (1 - \alpha) F_Z^{-1} \ge \alpha F_Y^{-1} + (1 - \alpha)F_Z^{-1}$. This leads to an alternate, dual, theory of choice than that of expected utility. Under these axioms the decision policy behaves as though it is maximizing a distorted expectation, for some continuous monotonic function $h$:
$$\pi(x) = \argmax_a \int_{-\infty}^{\infty} z \frac{\partial}{\partial z} (h \circ F_{Z(x, a)})(z) \,dz.$$

Such a function $h$ is known as a \textit{distortion risk measure}, as it distorts the cumulative probabilities of the random variable \cite{wang1996premium}. That is, 
we have two fundamentally equivalent approaches to risk-sensitivity.
Either, we choose a utility function and follow the expectation of this utility. Or, we choose a reweighting of the distribution and compute expectation under this distortion measure. Indeed, \citet{yaari1987dual} further showed that these two functions are inverses of each other. The choice between them amounts to a choice over whether the behavior should be invariant to mixing with random events or to convex combinations of outcomes.

Distortion risk measures include, as special cases, cumulative probability weighting used in cumulative prospect theory \cite{tversky1992advances}, conditional value at risk \cite{chow2014algorithms}, and many other methods \cite{morimura2010nonparametric}. Recently \citet{majumdar2017should} argued for the use of distortion risk measures in robotics.

\section{Implicit Quantile Networks}
\label{sec:analysis}

\begin{figure}
\begin{center}
\includegraphics[width=.48\textwidth]{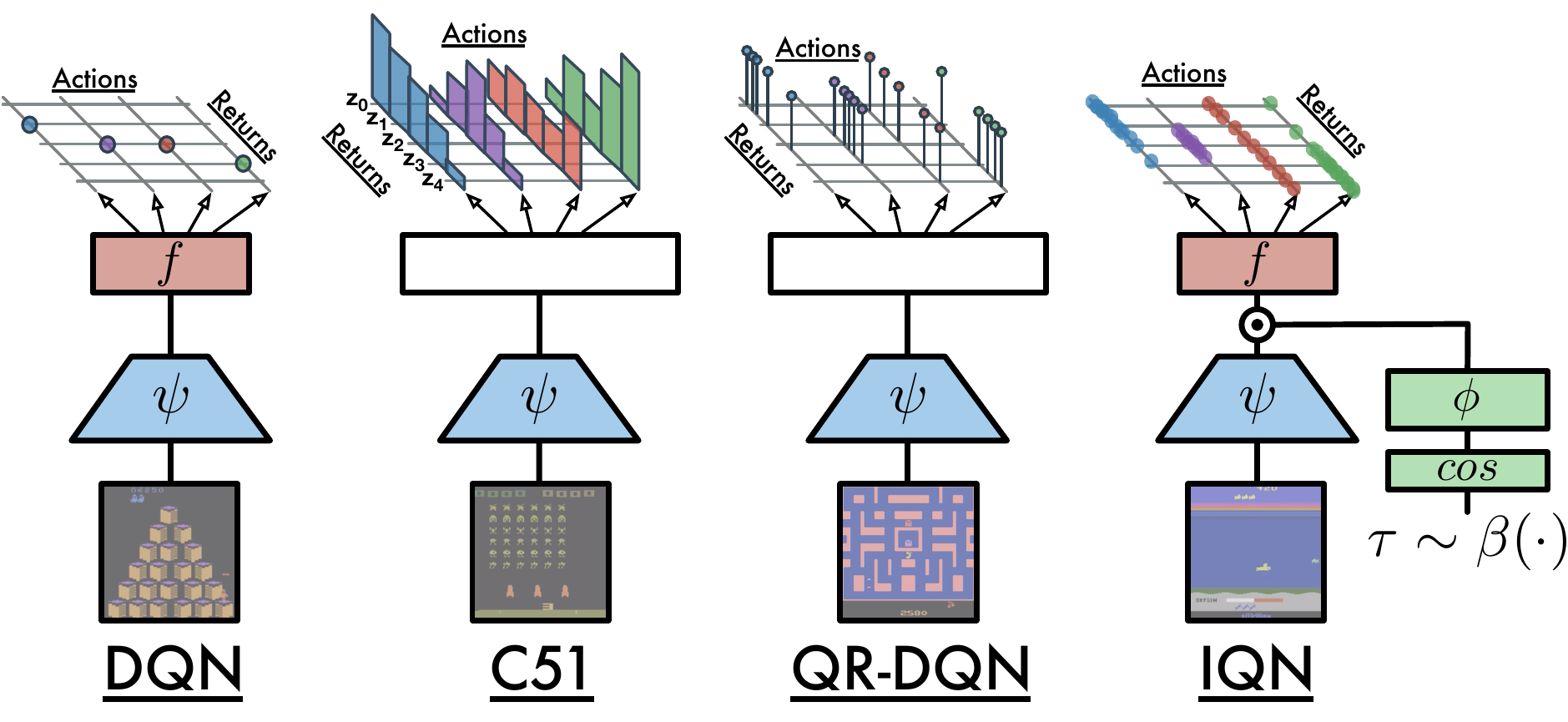}
\end{center}
\caption{Network architectures for DQN and recent distributional RL algorithms.}\label{fig:network_arch}
\end{figure}

We now introduce the \textit{implicit quantile network} (IQN), a deterministic parametric function trained to reparameterize samples from a base distribution, e.g.~$\tau \sim U([0,1])$, to the respective quantile values of a target distribution. 
IQN provides an effective way to learn an implicit representation of the return distribution, yielding a powerful function approximator for a new DQN-like agent.

Let $F^{-1}_Z(\tau)$ be the quantile function at $\tau \in [0, 1]$ for the random variable $Z$. For notational simplicity we write $Z_\tau := F^{-1}_Z(\tau)$, thus for $\tau \sim U([0,1])$ the resulting state-action return distribution sample is $Z_\tau(x, a) \sim Z(x, a)$.

We propose to model the state-action quantile function as a mapping from state-actions and samples from some base distribution, typically $\tau \sim U([0,1])$, to $Z_\tau(x, a)$, viewed as samples from the implicitly defined return distribution.

Let $\beta\colon [0, 1] \to [0, 1]$ be a distortion risk measure, with identity corresponding to risk-neutrality. Then, the \textit{distorted expectation} of $Z(x, a)$ under $\beta$ is given by
\begin{equation*}
    Q_\beta(x, a) := \expect_{\tau \sim U([0,1])} \left[ Z_{\beta(\tau)}(x, a) \right].
\end{equation*}
Notice that the distorted expectation is equal to the expected value of $F^{-1}_{Z(x,a)}$ weighted by $\beta$, that is, $Q_\beta = \int_0^1 F^{-1}_Z(\tau) d\beta(\tau)$. The immediate implication of this is that for any $\beta$, there exists a sampling distribution for $\tau$ such that the mean of $Z_\tau$ is equal to the distorted expectation of $Z$ under $\beta$, that is, any distorted expectation can be represented as a weighted sum over the quantiles \cite{dhaene2012remarks}. Denote by $\pi_\beta$ the risk-sensitive greedy policy
\begin{equation}\label{eqn:rs_policy}
    \pi_\beta(x) = \argmax_{a \in \cA} Q_\beta(x, a).
\end{equation}

For two samples $\tau, \tau' \sim U([0,1])$, and policy $\pi_\beta$, the sampled temporal difference (TD) error at step $t$ is
\begin{equation}\label{eqn:sampledTD}
    \delta^{\tau,\tau'}_t = r_t + \gamma Z_{\tau'}(x_{t+1}, \pi_\beta(x_{t+1})) - Z_{\tau}(x_t, a_t).
\end{equation}
Then, the IQN loss function is given by
\begin{equation}\label{eqn:iqn_loss}
    \cL(x_t, a_t, r_t, x_{t+1}) = \frac{1}{N'} \sum_{i=1}^{N} \sum_{j=1}^{N'} \rho_{\tau_i}^\kappa \left( \delta_t^{\tau_i, \tau_j'} \right),
\end{equation}
where $N$ and $N'$ denote the respective number of iid samples $\tau_i, \tau_j' \sim U([0,1])$ used to estimate the loss.
A corresponding sample-based risk-sensitive policy is obtained by approximating $Q_\beta$ in Equation~\ref{eqn:rs_policy} by $K$ samples of $\tilde \tau \sim U([0,1])$:
\begin{equation*}
    \tilde \pi_\beta(x) = \argmax_{a \in \cA} \frac{1}{K}\sum_{k=1}^K  Z_{\beta(\tilde\tau_k)}(x, a).
\end{equation*}

Implicit quantile networks differ from the approach of \citet{dabney2017qr} in two ways. First, instead of approximating the quantile function at $n$ fixed values of $\tau$ we approximate it with $Z_\tau(x, a) \approx f( \psi(x), \phi(\tau))_a$ for some differentiable functions $f$, $\psi$, and $\phi$. If we ignore the distributional interpretation for a moment and view each $Z_\tau(x, a)$ as a separate action-value function, this highlights that implicit quantile networks are a type of \textit{universal value function approximator} (UVFA) \cite{schaul2015universal}. There may be additional benefits to implicit quantile networks beyond the obvious increase in representational fidelity. As with UVFAs, we might hope that training over many different $\tau$'s (goals in the case of the UVFA) leads to better generalization between values and improved sample complexity than attempting to train each separately.

Second, $\tau$, $\tau'$, and $\tilde \tau$ are sampled from continuous, independent, distributions. Besides $U([0,1])$, we also explore risk-sentive policies $\pi_\beta$, with non-linear $\beta$. The independent sampling of each $\tau$, $\tau'$ results in the sample TD errors being decorrelated, and the estimated action-values go from being the true mean of a mixture of $n$ Diracs to a sample mean of the implicit distribution defined by reparameterizing the sampling distribution via the learned quantile function.

\subsection{Implementation}
\label{sec:algorithm}
Consider the neural network structure used by the DQN agent \cite{mnih15nature}. Let $\psi\colon \cX \to \bR^d$ be the function computed by the convolutional layers and $f\colon \bR^d \to \bR^{|\cA|}$ the subsequent fully-connected layers mapping $\psi(x)$ to the estimated action-values, such that $Q(x, a) \approx f(\psi(x))_a$. For our network we use the same functions $\psi$ and $f$ as in DQN, but include an additional function $\phi\colon [0, 1] \to \bR^d$ computing an embedding for the sample point $\tau$. We combine these to form the approximation $Z_\tau(x, a) \approx f(\psi(x) \odot \phi(\tau))_a$, where $\odot$ denotes the element-wise (Hadamard) product.

As the network for $f$ is not particularly deep, we use the multiplicative form, $\psi \odot \phi$, to force interaction between the convolutional features and the sample embedding. Alternative functional forms, e.g.~concatenation or a `residual' function $\psi \odot (1 + \phi)$, are conceivable, and $\phi(\tau)$ can be parameterized in different ways. To investigate these, we compared performance across a number of architectural variants on six Atari 2600 games (\textsc{Asterix, Assault, Breakout, Ms.Pacman, QBert, Space Invaders}).
Full results are given in the Appendix. Despite minor variation in performance, we found the general approach to be robust to the various choices. Based upon the results we used the following function in our later experiments, for embedding dimension $n = 64$:
\begin{equation}\label{eqn:iqn_architecture}
    \phi_j(\tau) := \operatorname{ReLU}(\sum_{i=0}^{n-1} \cos(\pi i \tau)w_{ij} + b_j).
\end{equation}

After settling on a network architecture, we study the effect of 
the number of samples, $N$ and $N'$, used in the estimate terms of Equation~\ref{eqn:iqn_loss}.

We hypothesized that $N$, the number of samples of $\tau \sim U([0,1])$, would affect the sample complexity of IQN, with larger values leading to faster learning, and that with $N = 1$ one would potentially approach the performance of DQN. This would support the hypothesis that the improved performance of many distributional RL algorithms rests on their effect as auxiliary loss functions, which would vanish in the case of $N = 1$.
Furthermore, we believed that $N'$, the number of samples of $\tau' \sim U([0,1])$, would affect the variance of the gradient estimates much like a mini-batch size hyperparameter. Our prediction was that $N'$ would have the greatest effect on variance of the long-term performance of the agent.

\begin{figure}
\begin{center}
\includegraphics[width=.5\textwidth]{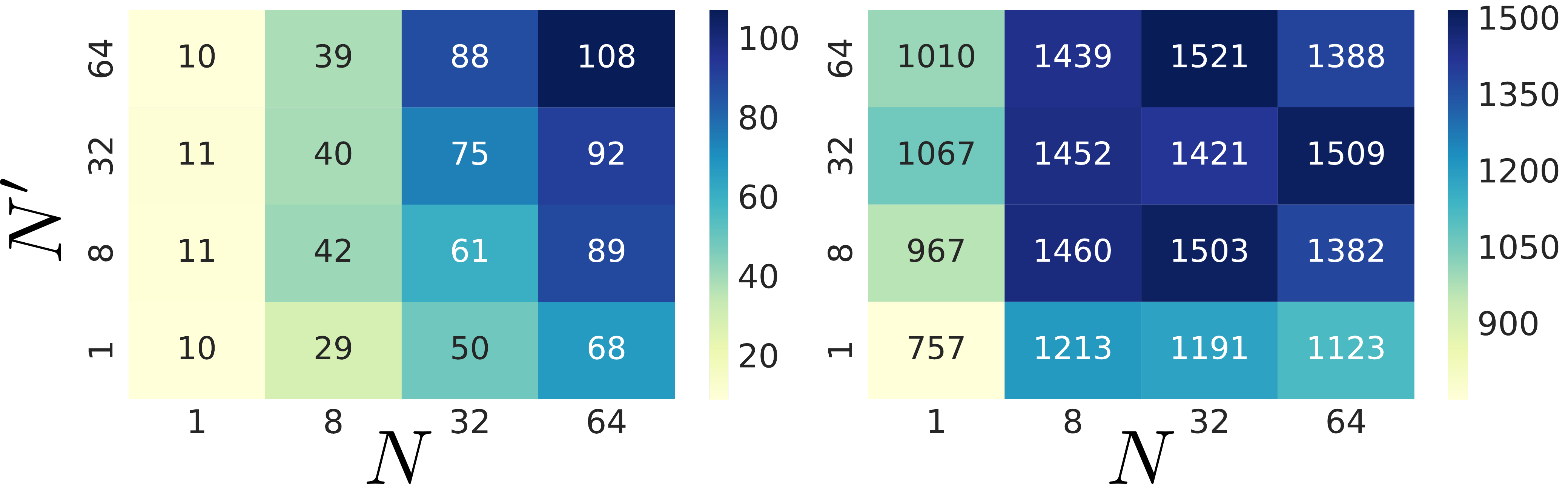}
\end{center}
\caption{Effect of varying $N$ and $N'$, the number of samples used in the loss function in Equation~\ref{eqn:iqn_loss}. Figures show human-normalized agent performance, averaged over six Atari games, averaged over first 10M frames of training (left) and last 10M frames of training (right). Corresponding values for baselines: DQN ($32, 253$) and QR-DQN ($144, 1243$).}\label{fig:num_atoms}
\end{figure}

We used the same set of six games as before, with our chosen architecture, and varied $N, N' \in \{1, 8, 32, 64\}$. In Figure~\ref{fig:num_atoms} we report the average human-normalized scores on the six games for each configuration. Figure~\ref{fig:num_atoms} (left) shows the average performance over the first ten million frames, while (right) shows the average performance over the last ten million (from 190M to 200M). 

As expected, we found that $N$ has a dramatic effect on early performance, shown by the continual improvement in score as the value increases.
Additionally, we observed that $N'$ affected performance very differently than expected: it had a strong effect on early performance, but minimal impact on long-term performance past $N' = 8$. 

Overall, while using more samples for both distributions is generally favorable, $N = N' = 8$ appears to be sufficient to achieve the majority of improvements offered by IQN for long-term performance, with variation past this point largely insignificant. To our surprise we found that even for $N = N' = 1$, which is comparable to DQN in the number of loss components, the longer term performance is still quite strong ($\approx3\times$ DQN).

In an informal evaluation, we did not find IQN to be sensitive to $K$, the number of samples used for the policy, and have fixed it at $K = 32$ for all experiments.

\section{Risk-Sensitive Reinforcement Learning}
\label{sec:risky_rl}

\begin{figure*}
\begin{center}
\includegraphics[width=\textwidth]{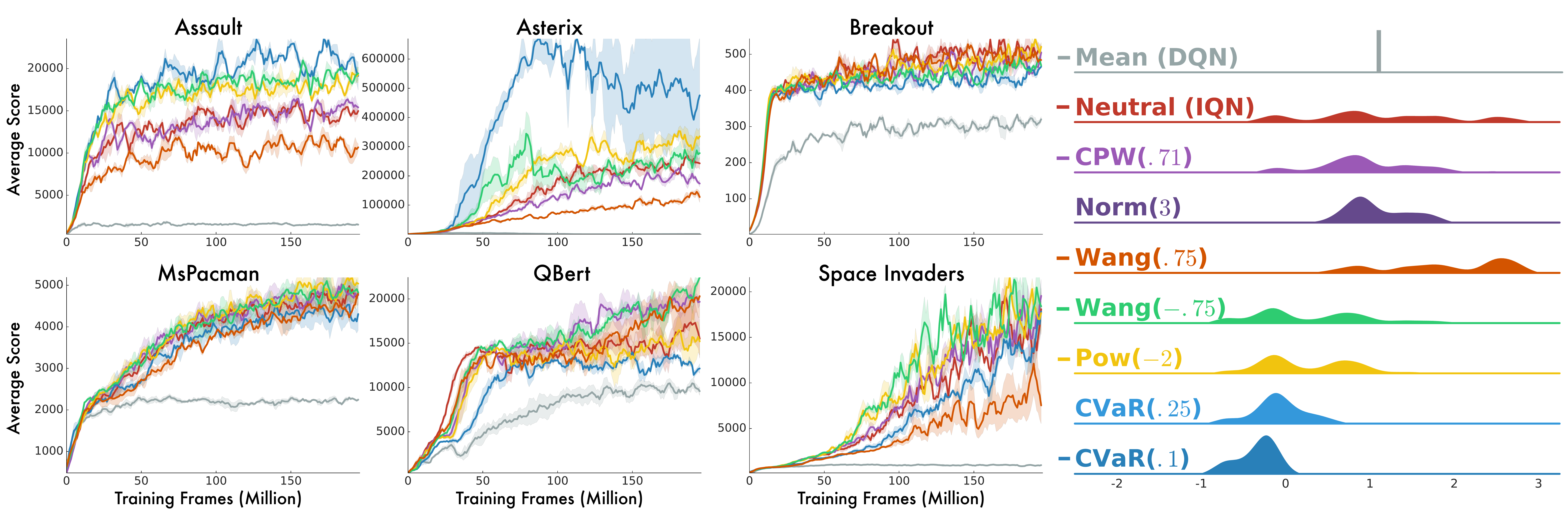}
\end{center}
\caption{Effects of various changes to the sampling distribution, that is various cumulative probability weightings.}\label{fig:risk_atari}
\end{figure*}

In this section, we explore the effects of varying the distortion risk measure, $\beta$, away from identity. This only affects the policy, $\pi_\beta$, used both in Equation~\ref{eqn:sampledTD} and for acting in the environment. As we have argued, evaluating under different distortion risk measures is equivalent to changing the sampling distribution for $\tau$, allowing us to achieve various forms of risk-sensitive policies. We focus on a handful of sampling distributions and their corresponding distortion measures. The first one is the cumulative probability weighting parameterization proposed in cumulative prospect theory \cite{tversky1992advances,gonzalez1999shape}:
\begin{equation*}
    \operatorname{CPW}(\eta, \tau) = \frac{\tau^{\eta}}{(\tau^{\eta} + (1 - \tau)^\eta)^{\frac{1}{\eta}}}.
\end{equation*}
In particular, we use the parameter value $\eta = 0.71$ found by \citet{wu1996curvature} to most closely match human subjects. This choice is interesting as, unlike the others we consider, it is neither globally convex nor concave. For small values of $\tau$ it is locally concave and for larger values of $\tau$ it becomes locally convex. Recall that concavity corresponds to risk-averse and convexity to risk-seeking policies.

Second, we consider the distortion risk measure proposed by \citet{wang2000class}, where $\Phi$ and $\Phi^{-1}$ are taken to be the standard Normal cumulative distribution function and its inverse:
\begin{equation*}
    \operatorname{Wang}(\eta, \tau) = \Phi(\Phi^{-1}(\tau) + \eta).
\end{equation*}
For $\eta < 0$, this produces risk-averse policies and we include it due to its simple interpretation and ability to switch between risk-averse and risk-seeking distortions.

Third, we consider a simple power formula for risk-averse ($\eta < 0$) or risk-seeking ($\eta > 0$) policies:
\begin{equation*}
    \operatorname{Pow}(\eta, \tau) =  \begin{cases}
        \tau^{\frac{1}{1 + |\eta|}},\quad \ &\text{if } \eta \ge 0\\
        1 - (1 - \tau)^{\frac{1}{1 + |\eta|}},\quad \ &\text{otherwise}
    \end{cases}.
\end{equation*}

Finally, we consider conditional value-at-risk (CVaR):
\begin{equation*}
    \operatorname{CVaR}(\eta, \tau) = \eta \tau.
\end{equation*}
CVaR has been widely studied in and out of reinforcement learning \cite{chow2014algorithms}. Its implementation as a modification to the sampling distribution of $\tau$ is particularly simple, as it changes $\tau \sim U([0,1])$ to $\tau \sim U([0,\eta])$. Another interesting sampling distribution, not included in our experiments, is denoted $\operatorname{Norm}(\eta)$ and corresponds to $\tau$ sampled by averaging $\eta$ samples from $U([0,1])$.

In Figure~\ref{fig:risk_atari} (right) we give an example of a distribution (Neutral) and how each of these distortion measures affects the implied distribution due to changing the sampling distribution of $\tau$. $\operatorname{Norm}(3)$ and $\operatorname{CPW}(.71)$ reduce the impact of the tails of the distribution, while $\operatorname{Wang}$ and $\operatorname{CVaR}$ heavily shift the distribution mass towards the tails, creating a risk-averse or risk-seeking preference. Additionally, while CVaR entirely ignores all values corresponding to $\tau > \eta$, $\operatorname{Wang}$ gives these non-zero, but vanishingly small, probability.

By using these sampling distributions we can induce various risk-sensitive policies in IQN. We evaluate these on the same set of six Atari 2600 games previously used. Our algorithm simply changes the policy to maximize the distorted expectations instead of the usual sample mean. Figure~\ref{fig:risk_atari} (left) shows our results in this experiment, with average scores reported under the usual, risk-neutral, evaluation criterion.

Intuitively, we expected to see a qualitative effect from risk-sensitive training, e.g.~strengthened exploration from a risk-seeking objective. Although we did see qualitative differences, these did not always match our expectations.
For two of the games, \textsc{Asterix} and \textsc{Assault}, there is a very significant advantage to the risk-averse policies. Although $\operatorname{CPW}$ tends to perform almost identically to the standard risk-neutral policy, and the risk-seeking $\operatorname{Wang}(1.5)$ performs as well or worse than risk-neutral, we find that both risk-averse policies improve performance over standard IQN. However, we also observe that the more risk-averse of the two, $\operatorname{CVaR}(0.1)$, suffers some loss in performance on two other games (\textsc{QBert} and \textsc{Space Invaders}). 

Additionally, we note that the risk-seeking policy significantly underperforms the risk-neutral policy on three of the six games. It remains an open question as to exactly why we see improved performance for risk-averse policies. There are many possible explanations for this phenomenon, e.g.~that risk-aversion encodes a heuristic to stay alive longer, which in many games is correlated with increased rewards.

\section{Full Atari-57 Results}
\label{sec:atari57}

\begin{figure*}
\begin{center}
\includegraphics[width=\textwidth]{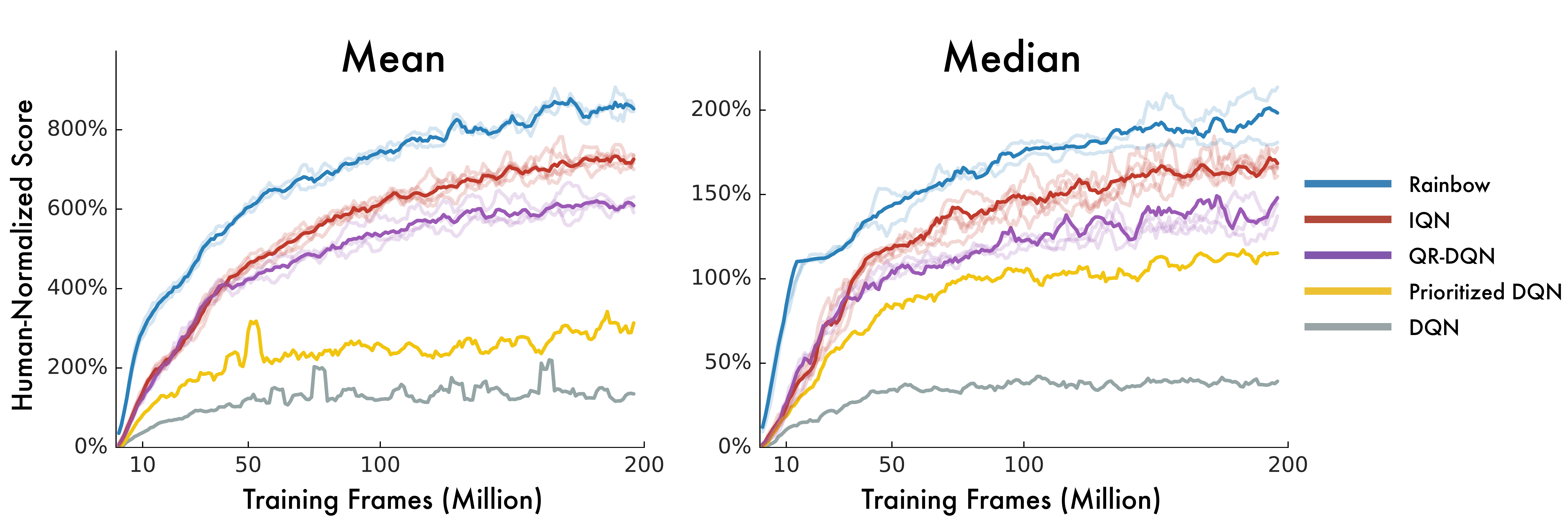}
\end{center}
\caption{Human-normalized mean (left) and median (right) scores on Atari-57 for IQN and various other algorithms. Random seeds shown as traces, with IQN averaged over 5, QR-DQN over 3, and Rainbow over 2 random seeds.}\label{fig:atari57}
\end{figure*}

Finally, we evaluate IQN on the full Atari-57 benchmark, comparing with the state-of-the-art performance of Rainbow, a distributional RL agent that combines several advances in deep RL \cite{hessel2018rainbow}, the closely related algorithm QR-DQN \cite{dabney2017qr}, prioritized experience replay DQN \cite{schaul16prioritized}, and the original DQN agent \cite{mnih15nature}. Note that in this section we use the risk-neutral variant of the IQN, that is, the policy of the IQN agent is the regular $\epsilon$-greedy policy with respect to the mean of the state-action return distribution.

It is important to remember that Rainbow builds upon the distributional RL algorithm C51 \cite{c51}, but also includes prioritized experience replay \cite{schaul16prioritized}, Double DQN \cite{vanhasselt16deep}, Dueling Network architecture \cite{wang2016dueling}, Noisy Networks \cite{fortunato2017noisy}, and multi-step updates \cite{sutton1988learning}. In particular, besides the distributional update, $n$-step updates and prioritized experience replay were found to have significant impact on the performance of Rainbow. Our other competitive baseline is QR-DQN, which is currently state-of-the-art for agents that do not combine distributional updates, $n$-step updates, and prioritized replay.

Thus, between QR-DQN and the much more complex Rainbow we compare to the two most closely related, and best performing, agents in published work. In particular, we would expect that IQN would benefit from the additional enhancements in Rainbow, just as Rainbow improved significantly over C51.

Figure~\ref{fig:atari57} shows the mean (left) and median (right) human-normalized scores during training over the Atari-57 benchmark. IQN dramatically improves over QR-DQN, which itself improves on many previously published results. At 100 million frames IQN has reached the same level of performance as QR-DQN at 200 million frames. Table~\ref{fig:perc_scores} gives a comparison between the same methods in terms of their best, human-normalized, scores per game under the 30 random no-op start condition. These are averages over the given number of seeds. Additionally, using human-starts, IQN achieves $162\%$ median human-normalized score, whereas Rainbow reaches $153\%$ \cite{hessel2018rainbow}, see Table~\ref{fig:perc_scores_human}.

\begin{table}[ht]
\begin{center}
\begin{tabular}{ l | r | r | r | c }
\multicolumn{1}{c}{} & \mbox{\textbf{Mean}} & \mbox{\textbf{Median}} & \mbox{\textbf{Human Gap}} & \mbox{\textbf{Seeds}} \\
\hline
\textsc{DQN}  &   228\% & 79\% & 0.334 & 1 \\
\textsc{Prior.}   &   434\% & 124\% & 0.178  & 1 \\
\textsc{C51}   &   701\% & 178\% & 0.152  & 1 \\
\textsc{Rainbow}   &   \textbf{\textcolor{blue}{1189\%}} & \textbf{\textcolor{blue}{230\%}} & 0.144  & 2 \\
\textsc{QR-DQN}   &   864\% & 193\% & 0.165  & 3  \\
\hline
\textsc{IQN}   &   1019\% & 218\% & \textbf{\textcolor{blue}{0.141}} & 5  \\
\end{tabular}
\end{center}
\caption{Mean and median of scores across 57 Atari 2600 games, measured as percentages of human baseline \cite{nair15massively}. Scores are averages over number of seeds.}
\label{fig:perc_scores}
\end{table}

\begin{table}[ht]
\begin{center}
\begin{tabular}{cccccc}
\multicolumn{6}{c}{\textbf{Human-starts (median)}}  \\
\hline
\textsc{DQN}  & \textsc{Prior.} & \textsc{A3C}   & \textsc{C51}   & \textsc{Rainbow} & \textsc{IQN}   \\
\hline
68\% & 128\%  & 116\% & 125\% & 153\%   & \textbf{\textcolor{blue}{162\%}}
\end{tabular}
\end{center}
\caption{Median human-normalized scores for human-starts.}
\label{fig:perc_scores_human}
\end{table}

Finally, we took a closer look at the games in which each algorithm continues to underperform humans, and computed, on average, how far below human-level they perform\footnote{Details of how this is computed can be found in the Appendix.}. We refer to this value as the \textit{human-gap}\footnote{Thanks to Joseph Modayil for proposing this metric.} metric and give results in Table~\ref{fig:perc_scores}. Interestingly, C51 outperforms QR-DQN in this metric, and IQN outperforms all others. This shows that the remaining gap between Rainbow and IQN is entirely from games on which both algorithms are already super-human. The games where the most progress in RL is needed happen to be the games where IQN shows the greatest improvement over QR-DQN and Rainbow.

\section{Discussion and Conclusions}
\label{sec:discussion}

We have proposed a generalization of recent work based around using quantile regression to learn the distribution over returns of the current policy. Our generalization leads to a simple change to the DQN agent to enable distributional RL, the natural integration of risk-sensitive policies, and significantly improved performance over existing methods. The IQN algorithm provides, for the first time, a fully integrated distributional RL agent without prior assumptions on the parameterization of the return distribution.

IQN can be trained with as little as a single sample from each state-action value distribution, or as many as computational limits allow to improve the algorithm's data efficiency. Furthermore, IQN allows us to expand the class of control policies to a large class of risk-sensitive policies connected to distortion risk measures. Finally, we show substantial gains on the Atari-57 benchmark over QR-DQN, and even halving the distance between QR-DQN and Rainbow.

Despite the significant empirical successes in this paper there are many areas in need of additional theoretical analysis. We highlight a few particularly relevant open questions we were unable to address in the present work. First, sample-based convergence results have been recently shown for a class of categorical distributional RL algorithms \cite{rowland2018analysis}. Could existing sample-based RL convergence results be extended to the QR-based algorithms?

Second, can the contraction mapping results for a fixed grid of quantiles given by \citet{dabney2017qr} be extended to the more general class of approximate quantile functions studied in this work? Finally, and particularly salient to our experiments with distortion risk measures, theoretical guarantees for risk-sensitive RL have been building over recent years, but have been largely limited to special cases and restricted classes of risk-sensitive policies. Can the convergence of the distribution of returns under the Bellman operator be leveraged to show convergence to a fixed-point in distorted expectations? In particular, can the control results of \citet{c51} be expanded to cover some class of risk-sensitive policies?

There remain many intriguing directions for future research into distributional RL, even on purely empirical fronts. \citet{hessel2018rainbow} recently showed that distributional RL agents can be significantly improved, when combined with other techniques. Creating a Rainbow-IQN agent could yield even greater improvements on Atari-57. We also recall the surprisingly rich return distributions found by \citet{barthmaron2018d4pg}, and hypothesize that the continuous control setting may be a particularly fruitful area for the application of distributional RL in general, and IQN in particular.


\clearpage

\bibliography{distrl}
\bibliographystyle{icml2018}

\clearpage

\section*{Appendix}
\begin{figure}[b!]
\begin{center}
\includegraphics[width=\textwidth]{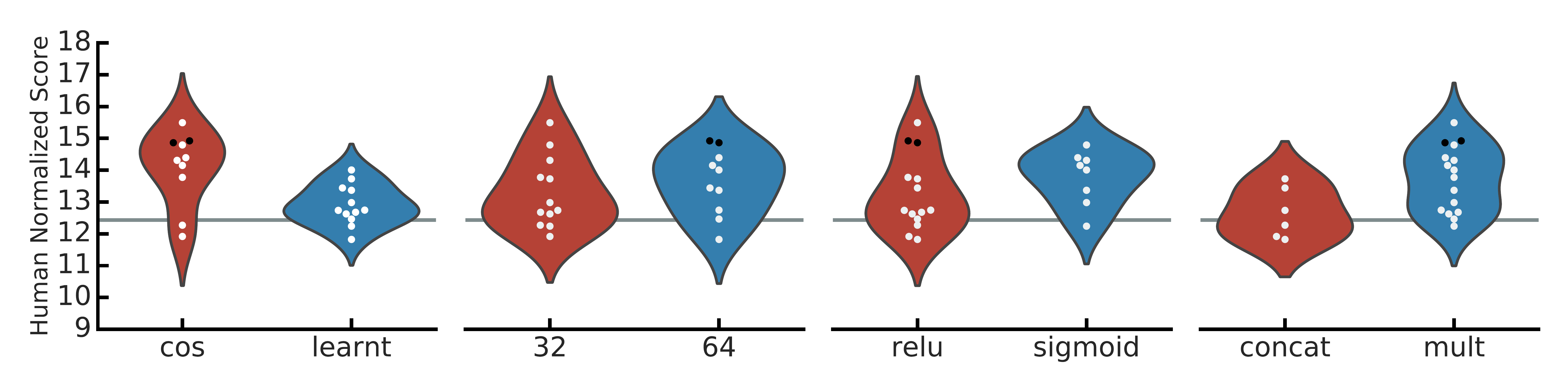}
\end{center}
\caption{Comparison of architectural variants.}
\label{fig:hyper_comparisons}
\end{figure}

\subsection*{Architecture and Hyperparameters}
\label{sec:hyper}

We considered multiple architectural variants for parameterizing an IQN. All of these build on the Q-network of a regular DQN \cite{mnih15nature}, which can be seen as the composition of a convolutional stack $\psi \colon \cX \to \bR^d$ and an MLP $f\colon \bR^d \to \bR^{|\cA|}$, and extend it by an embedding of the sample point, $\phi\colon [0,1] \to \bR^d$, and a merging function $m \colon \bR^d \times \bR^d \to \bR^d$, resulting in the function
\begin{equation*}
    \operatorname{IQN}(x, \tau) = f(m(\psi(x), \phi(\tau))).
\end{equation*}

For the embedding $\phi$, we considered a number of variants: a learned linear embedding, a learned MLP embedding with a single hidden layer of size $n$, and a learned linear function of $n$ cosine basis functions of the form $\cos(\pi i \tau), i = 1, \dots, n$. Each of those was followed by either a ReLU or sigmoid nonlinearity. 

For the merging function $m$, the simplest choice would be a simple vector concatenation of $\psi(x)$ and $\phi(\tau)$. Note however, that the MLP $f$ which takes in the output of $m$ and outputs the action-value quantiles, only has a single hidden layer in the DQN network. Therefore, to force a sufficiently early interaction between the two representations, we also considered a multiplicative function $m(\psi, \phi) = \psi \odot \phi$, where $\odot$ denotes the element-wise (Hadamard) product of two vectors, as well as a `residual' function $m(\psi, \phi) = \psi \odot (1 + \phi)$.

Early experiments showed that a simple linear embedding of $\tau$ was insufficient to achieve good performance, and the residual version of $m$ didn't show any marked difference to the multiplicative variant, so we do not include results for these here. For the other configurations, Figure~\ref{fig:hyper_comparisons} shows pairwise comparisons between 1) a cosine basis function embedding and a completely learned MLP embedding, 2) an embedding size (hidden layer size or number of cosine basis elements) 32 and 64, 3) ReLU and sigmoid nonlinearity following the embedding, and 4) concatenation and a multiplicative interaction between $\psi(x)$ and $\phi(\tau)$. 

Each comparison `violin plot' can be understood as a marginalization over the other variants of the architecture, with the human-normalized performance at the end of training, averaged across six Atari 2600 games, on the y-axis. Each white dot corresponds to a configuration (each represented by two seeds), the black dots show the position of our preferred configuration. The width of the colored regions corresponds to a kernel density estimate of the number of configurations at each performance level. 

Our final choice is a multiplicative interaction with a linear function of a cosine embedding, with $n=64$ and a ReLU nonlinearity (see Equation~\ref{eqn:iqn_architecture}), as this configuration yielded the highest performance consistently over multiple seeds. Also noteworthy is the overall robustness of the approach to these variations: most of the configurations consistently outperform the QR-DQN baseline shown as a grey horizontal line for comparison.

We give pseudo-code for the IQN loss in Algorithm~\ref{alg:iqn}.
All other hyperparameters for this agent correspond to the ones used by \citet{dabney2017qr}. In particular, the Bellman target is computed using a target network. Notice that IQN will generally be more computationally expensive per-sample than QR-DQN. However, in practice IQN requires many fewer samples per update than QR-DQN so that the actual running times are comparable.

\begin{algorithm}[ht]
\caption{Implicit Quantile Network Loss}\label{alg:iqn}
\begin{algorithmic}
\REQUIRE $N, N', K, \kappa$ and functions $\beta, Z$
\INPUT $x, a, r, x'$, $\gamma \in [0, 1)$
\STATE \textcolor{gray}{\# Compute greedy next action}
\STATE $a^* \leftarrow \argmax_{a'} \frac{1}{K} \sum_{k}^K Z_{\tilde \tau_k}(x', a'),\ \quad \tilde\tau_k \sim \beta(\cdot)$
\STATE \textcolor{gray}{\# Sample quantile thresholds}
\STATE $\tau_i, \tau'_j \sim U([0, 1]),\ \quad 1 \le i \le N, 1 \le j \le N'$
\STATE \textcolor{gray}{\# Compute distributional temporal differences}
\STATE $\delta_{ij} \leftarrow r + \gamma Z_{\tau'_j}(x', a^*) - Z_{\tau_i}(x, a),\ \quad \forall i,j$
\STATE \textcolor{gray}{\# Compute Huber quantile loss}
\OUTPUT $\sum_{i=1}^{N} \expect_{\tau'} \left[ \rho^\kappa_{\tau_i}( \delta_{ij}) \right]$
\end{algorithmic}
\end{algorithm}

\newpage
\subsection*{Evaluation}
\label{sec:eval}

The human-normalized scores reported in this paper are given by the formula \cite{vanhasselt16deep, dabney2017qr}
\begin{equation*}
    score = \frac{agent - random}{human - random},
\end{equation*}
where $agent$, $human$ and $random$ are the per-game raw scores (undiscounted returns) for the given agent, a reference human player, and random agent baseline \cite{mnih15nature}.

The `human-gap' metric referred to at the end of Section~\ref{sec:atari57} builds on the human-normalized score, but emphasizes the remaining improvement for the agent to reach super-human performance. It is given by $gap = \max(1 - score, 0)$,
with a value of $1$ corresponding to random play, and a value of $0$ corresponding to super-human level of performance. To avoid degeneracies in the case of $human < random$, the quantity is being clipped above at $1$.

\begin{figure*}
\begin{center}
\includegraphics[width=\textwidth]{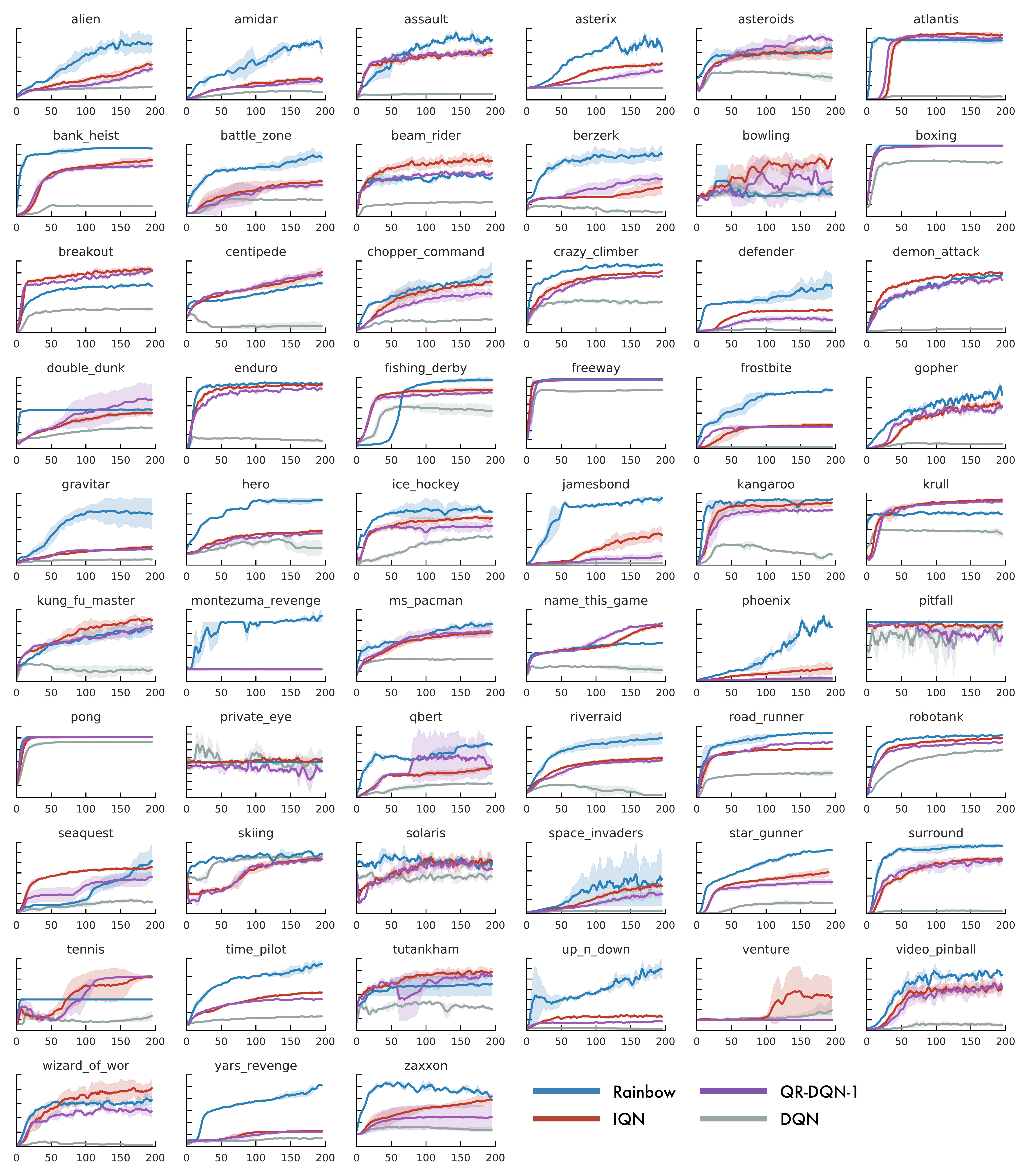}
\end{center}
\caption{Complete Atari-57 training curves.}
\end{figure*}

\newpage

\begin{figure*}
\small
\centering
\begin{tabular}{ l | r|r|r|r|r|r}
  \textbf{\textsc{games}}  &  \textbf{\textsc{random}}  &  \textbf{\textsc{human}}  &  \textbf{\textsc{dqn}}  &   \textbf{\textsc{prior.}} \textbf{\textsc{duel.}}  &  \textbf{\textsc{qr-dqn}} &  \textbf{\textsc{iqn}}\\
\hline
Alien & 227.8 & 7,127.7 & 1,620.0 & 3,941.0 & 4,871 & \textbf{\textcolor{blue}{7,022}} \\
Amidar & 5.8 & 1,719.5 & 978.0 & 2,296.8 & 1,641 & \textbf{\textcolor{blue}{2,946}} \\
Assault & 222.4 & 742.0 & 4,280.4 & 11,477.0 & 22,012 & \textbf{\textcolor{blue}{29,091}} \\
Asterix & 210.0 & 8,503.3 & 4,359.0 & \textbf{\textcolor{blue}{375,080.0}} & 261,025 & 342,016 \\
Asteroids & 719.1 & 47,388.7 & 1,364.5 & 1,192.7 & \textbf{\textcolor{blue}{4,226}} & 2,898 \\
Atlantis & 12,850.0 & 29,028.1 & 279,987.0 & 395,762.0 & 971,850 & \textbf{\textcolor{blue}{978,200}} \\
Bank Heist & 14.2 & 753.1 & 455.0 & \textbf{\textcolor{blue}{1,503.1}} & 1,249 & 1,416 \\
Battle Zone & 2,360.0 & 37,187.5 & 29,900.0 & 35,520.0 & 39,268 & \textbf{\textcolor{blue}{42,244}} \\
Beam Rider & 363.9 & 16,926.5 & 8,627.5 & 30,276.5 & 34,821 & \textbf{\textcolor{blue}{42,776}} \\
Berzerk & 123.7 & 2,630.4 & 585.6 & \textbf{\textcolor{blue}{3,409.0}} & 3,117 & 1,053 \\
Bowling & 23.1 & 160.7 & 50.4 & 46.7 & 77.2 & \textbf{\textcolor{blue}{86.5}} \\
Boxing & 0.1 & 12.1 & 88.0 & 98.9 & \textbf{\textcolor{blue}{99.9}} & 99.8 \\
Breakout & 1.7 & 30.5 & 385.5 & 366.0 & \textbf{\textcolor{blue}{742}} & 734 \\
Centipede & 2,090.9 & 12,017.0 & 4,657.7 & 7,687.5 & \textbf{\textcolor{blue}{12,447}} & 11,561 \\
Chopper Command & 811.0 & 7,387.8 & 6,126.0 & 13,185.0 & 14,667 & \textbf{\textcolor{blue}{16,836}} \\
Crazy Climber & 10,780.5 & 35,829.4 & 110,763.0 & 162,224.0 & 161,196 & \textbf{\textcolor{blue}{179,082}} \\
Defender & 2,874.5 & 18,688.9 & 23,633.0 & 41,324.5 & 47,887 & \textbf{\textcolor{blue}{53,537}} \\
Demon Attack & 152.1 & 1,971.0 & 12,149.4 & 72,878.6 & 121,551 & \textbf{\textcolor{blue}{128,580}} \\
Double Dunk & -18.6 & -16.4 & -6.6 & -12.5 & \textbf{\textcolor{blue}{21.9}} & 5.6 \\
Enduro & 0.0 & 860.5 & 729.0 & 2,306.4 & 2,355 & \textbf{\textcolor{blue}{2,359}} \\
Fishing Derby & -91.7 & -38.7 & -4.9 & \textbf{\textcolor{blue}{41.3}} & 39.0 & 33.8 \\
Freeway & 0.0 & 29.6 & 30.8 & 33.0 & \textbf{\textcolor{blue}{34.0}} & \textbf{\textcolor{blue}{34.0}} \\
Frostbite & 65.2 & 4,334.7 & 797.4 & \textbf{\textcolor{blue}{7,413.0}} & 4,384 & 4,324 \\
Gopher & 257.6 & 2,412.5 & 8,777.4 & 104,368.2 & 113,585 & \textbf{\textcolor{blue}{118,365}} \\
Gravitar & 173.0 & 3,351.4 & 473.0 & 238.0 & \textbf{\textcolor{blue}{995}} & 911 \\
H.E.R.O. & 1,027.0 & 30,826.4 & 20,437.8 & 21,036.5 & 21,395 & \textbf{\textcolor{blue}{28,386}} \\
Ice Hockey & -11.2 & 0.9 & -1.9 & -0.4 & -1.7 & \textbf{\textcolor{blue}{0.2}} \\
James Bond & 29.0 & 302.8 & 768.5 & 812.0 & 4,703 & \textbf{\textcolor{blue}{35,108}} \\
Kangaroo & 52.0 & 3,035.0 & 7,259.0 & 1,792.0 & 15,356 & \textbf{\textcolor{blue}{15,487}} \\
Krull & 1,598.0 & 2,665.5 & 8,422.3 & 10,374.4 & \textbf{\textcolor{blue}{11,447}} & 10,707 \\
Kung-Fu Master & 258.5 & 22,736.3 & 26,059.0 & 48,375.0 & \textbf{\textcolor{blue}{76,642}} & 73,512 \\
Montezuma’s Revenge & 0.0 & 4,753.3 & \textbf{\textcolor{blue}{0.0}} & \textbf{\textcolor{blue}{0.0}} & \textbf{\textcolor{blue}{0.0}} & \textbf{\textcolor{blue}{0.0}} \\
Ms. Pac-Man & 307.3 & 6,951.6 & 3,085.6 & 3,327.3 & 5,821 & \textbf{\textcolor{blue}{6,349}} \\
Name This Game & 2,292.3 & 8,049.0 & 8,207.8 & 15,572.5 & 21,890 & \textbf{\textcolor{blue}{22,682}} \\
Phoenix & 761.4 & 7,242.6 & 8,485.2 & \textbf{\textcolor{blue}{70,324.3}} & 16,585 & 56,599 \\
Pitfall! & -229.4 & 6,463.7 & -286.1 & \textbf{\textcolor{blue}{0.0}} & \textbf{\textcolor{blue}{0.0}} & \textbf{\textcolor{blue}{0.0}} \\
Pong & -20.7 & 14.6 & 19.5 & 20.9 & \textbf{\textcolor{blue}{21.0}} & \textbf{\textcolor{blue}{21.0}} \\
Private Eye & 24.9 & 69,571.3 & 146.7 & 206.0 & \textbf{\textcolor{blue}{350}} & 200 \\
Q*Bert & 163.9 & 13,455.0 & 13,117.3 & 18,760.3 & \textbf{\textcolor{blue}{572,510}} & 25,750 \\
River Raid & 1,338.5 & 17,118.0 & 7,377.6 & \textbf{\textcolor{blue}{20,607.6}} & 17,571 & 17,765 \\
Road Runner & 11.5 & 7,845.0 & 39,544.0 & 62,151.0 & \textbf{\textcolor{blue}{64,262}} & 57,900 \\
Robotank & 2.2 & 11.9 & \textbf{\textcolor{blue}{63.9}} & 27.5 & 59.4 & 62.5 \\
Seaquest & 68.4 & 42,054.7 & 5,860.6 & 931.6 & 8,268 & \textbf{\textcolor{blue}{30,140}} \\
Skiing & -17,098.1 & -4,336.9 & -13,062.3 & -19,949.9 & -9,324 & \textbf{\textcolor{blue}{-9,289}} \\
Solaris & 1,236.3 & 12,326.7 & 3,482.8 & 133.4 & 6,740 & \textbf{\textcolor{blue}{8,007}} \\
Space Invaders & 148.0 & 1,668.7 & 1,692.3 & 15,311.5 & 20,972 & \textbf{\textcolor{blue}{28,888}} \\
Star Gunner & 664.0 & 10,250.0 & 54,282.0 & \textbf{\textcolor{blue}{125,117.0}} & 77,495 & 74,677 \\
Surround & -10.0 & 6.5 & -5.6 & 1.2 & 8.2 & \textbf{\textcolor{blue}{9.4}} \\
Tennis & -23.8 & -8.3 & 12.2 & 0.0 & \textbf{\textcolor{blue}{23.6}} & \textbf{\textcolor{blue}{23.6}} \\
Time Pilot & 3,568.0 & 5,229.2 & 4,870.0 & 7,553.0 & 10,345 & \textbf{\textcolor{blue}{12,236}} \\
Tutankham & 11.4 & 167.6 & 68.1 & 245.9 & \textbf{\textcolor{blue}{297}} & 293 \\
Up and Down & 533.4 & 11,693.2 & 9,989.9 & 33,879.1 & 71,260 & \textbf{\textcolor{blue}{88,148}} \\
Venture & 0.0 & 1,187.5 & 163.0 & 48.0 & 43.9 & \textbf{\textcolor{blue}{1,318}} \\
Video Pinball & 16,256.9 & 17,667.9 & 196,760.4 & 479,197.0 & \textbf{\textcolor{blue}{705,662}} & 698,045 \\
Wizard Of Wor & 563.5 & 4,756.5 & 2,704.0 & 12,352.0 & 25,061 & \textbf{\textcolor{blue}{31,190}} \\
Yars’ Revenge & 3,092.9 & 54,576.9 & 18,098.9 & \textbf{\textcolor{blue}{69,618.1}} & 26,447 & 28,379 \\
Zaxxon & 32.5 & 9,173.3 & 5,363.0 & 13,886.0 & 13,112 & \textbf{\textcolor{blue}{21,772}}
\end{tabular}
\caption{Raw scores for a single seed across all games, starting with 30 no-op actions. Reference values from \cite{wang2016dueling}.\label{fig:atari_sota}}
\end{figure*}

\end{document}